# DeepCPG Policies for Robot Locomotion

Aditya M. Deshpande, Eric Hurd, Ali A. Minai, and Manish Kumar

*Abstract*—Central Pattern Generators (CPGs) form the neural basis of the observed rhythmic behaviors for locomotion in legged animals. The CPG dynamics organized into networks allow the emergence of complex locomotor behaviors. In this work, we take this inspiration for developing walking behaviors in multi-legged robots. We present novel DeepCPG policies that embed CPGs as a layer in a larger neural network and facilitate end-to-end learning of locomotion behaviors in deep reinforcement learning (DRL) setup. We demonstrate the effectiveness of this approach on physics engine-based insectoid robots. We show that, compared to traditional approaches, DeepCPG policies allow sample-efficient end-to-end learning of effective locomotion strategies even in the case of high-dimensional sensor spaces (vision). We scale the DeepCPG policies using a modular robot configuration and multi-agent DRL. Our results suggest that gradual complexification with embedded priors of these policies in a modular fashion could achieve non-trivial sensor and motor integration on a robot platform. These results also indicate the efficacy of bootstrapping more complex intelligent systems from simpler ones based on biological principles. Finally, we present the experimental results for a proof-of-concept insectoid robot system for which DeepCPG learned policies initially using the simulation engine and these were afterwards transferred to real-world robots without any additional fine-tuning.

## I. INTRODUCTION

Biologically inspired robots often have many degrees of freedom. Locomotion in such legged robots involves the interaction of a multi-body system with the surrounding environments through multiple contact points. This presents a formidable challenge for traditional control approaches. Self-organization of complex behaviors is seen as a much more viable approach for these. While it may be possible to solve some of the challenges of autonomous locomotion in controlled environments, traditional approaches are not useful for the real world situations where multi-task generalization of the system is required.

Biologically, it is known that Central Pattern Generators (CPGs) are the neural modules primarily responsible for generating rhythmic responses that result in oscillatory functions [1]–[4]. CPGs are used in a number of biological functions like walking, swimming, flying, etc. [5] and have inspired development of elegant biomimetic control approaches for locomotion of legged robots [6]–[8]. Furthermore, sensory feedback also plays an important role in regulating the oscillatory behaviors of CPGs [9]. It has been reported that CPGs, sensory information and descending brain inputs interact with each other to orchestrate coordinated movement of the

A. M. Deshpande, E. Hurd, A. A. Minai and M. Kumar were with University of Cincinnati, Cincinnati, Ohio, 45221 USA e-mail: deshpaad@mail.uc.edu; hurdeg@mail.uc.edu; ali.minai@uc.edu; manish.kumar@uc.edu.

six legs of a walking insect [10], [11]. DeAngelis et al. [12] observed that sensory perturbations in walking Drosophila are responsible for altering their periodic walking gaits. Their findings suggested the variablity in Drosophila walking gaits could be a result of low-dimensional control architecture, which provides a framework for understanding the neural circuits that regulate hexapod legged locomotion

CPG networks are capable of generative encoding, and have inherent flexibility in combining phase coupling with traditional kinematic control to produce a variety of coordinated behaviors [13]–[18]. The incorporation of sensory feedback into CPGs has also been investigated, but this often requires extensive engineering. Thus, such methods have been developed for controlled scenarios including salamander-inspired Pleurobot [15], quadruped robots [19], worm-like robots [20], stick-insect robots [21], and dung beetle-like robot [22].

With advances in robotics, deep learning, and neuroscience, we are starting to see real-world robots that not only look like but also interact with the environment just as living creatures do [15], [23]–[26]. However, most of these need hand-tuned parameter sets and are only evaluated in the constrained or supervised setting. The models used in these robots are simplified further with various assumptions to reduce the number of tunable parameters. Although such methods have beautifully demonstrated how higher-level neural modulations in cortical centers could enable the emergence of various locomotion strategies [15], there has been limited work on how to extend these models for actively using high-dimensional complex observations from various on-board sensors to modulate the cortical signals.

Previous studies have demonstrated the usefulness of biomimetic movement templates such as movement primitives and CPGs in robots [27]–[34]. The Hodgkin-Huxley model of action potential generation in single neurons [35], [36] has been used for developing locomotion strategies in snake-like [37], [38] and quadruped robots [39]. The Matsuoka model [40] has been used to produce robot locomotion [17], [41]. The Van der Pol oscillator [42] and the Hopf oscillator [43] were used in swinging robots [44] and for generating walking gaits [45], [46]. The Kuramoto model of coupled phase oscillators is one of the most abstract and simple phenomenological models [47], [48] and is widely used in the robotics community to develop locomotion strategies [15], [17], [49], [50]. Most of these approaches have been developed with a behavior-specific objective for robots and employed parameter tuning for behavioral policies either using manual effort or auto-tuning with a genetic algorithm. For high-dimensional sensor and action spaces, deep learning methods embedded with movement primitive have been investigated in the context of imitation learning or supervised learning [51], [52]. Model-based control guided imitation learning has also been proposed for learning various quadruped robot gaits [53]. The end-to-end learning of behavioral control policies based on high-dimensional sensor feedback along with movement priors has received limited attention.

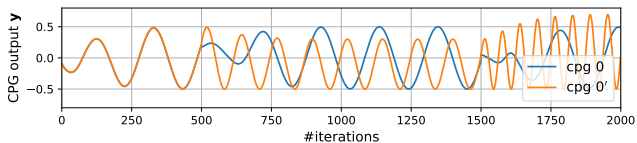

Fig. 1. Kuramoto model response for state and parameter perturbations: CPG 0 and CPG $0'$ are initialized at identical states. Plot contains CPG outputs $y_0$ and $y_{0'}$. (cpg 0) CPG states $\phi_0, a_0, b_0$ were perturbed to random values at iterations 500 and 1500. (cpg $0'$) CPG parameters $\omega_{0'}, A_{0'}, B_{0'}$ were perturbed to random values at iterations 500 and 1500. Refer equations (3)-(6) for details.

In this work, we focus on the problem of generality in controlling legged robots with high-dimensional environment sensing (observation space) and interaction modalities (action space) using CPG-inspired movement priors. We propose to bring together current ideas in reinforcement learning and deep learning, and expressive parameterizations resulting from CPGs in developing novel models that are capable of learning expressive behaviors for walking robots.

Brooks et al. [54] presented one of the first works that demonstrated feedback and reactive control laws alone can generate the full locomotive behaviors. Reinforcement learning (RL), on the other hand, offers a framework for artificial agents to *learn* such decision-making in their environments by trial and error using the feedback from their experiences [55]. The paradigm of modern deep reinforcement learning (DRL) algorithms has emerged with promising approaches to teach robots complex end-to-end control tasks such as locomotion [56], [57]. Typically, an RL agent (i.e. control policy) is trained to predict in raw action spaces and outputs actions in terms of motor torques or joint angles that may lead to non-smooth trajectories. It has been suggested previously by various works [58]–[60] that if we model the policy to predict actions in the trajectory space of the system, its response could be constrained to remain smooth. For example, in the case of a robotic system, if its control policy predicts the actions as trajectories of its various motor joints, the system response remains smooth even when these actions change (see cpg $0'$ in Fig. 1). We propose to use this principle in training the locomotion policies for the walking robots.

Various DRL algorithms have been proposed for learning legged robot locomotion where the control policies are trained from scratch. The work in [61] presented an attention-based recurrent encoder network for learning robust quadruped walking in challenging terrain. Authors in [62] presented a meta-RL approach that could adapt to different quadruped designs. Using locomotion priors with DRL has also been investigated in the past. Structured control networks proposed separation of linear and nonlinear parts of the control policy and using sinusoidal activations for the nonlinear part of the policy while training the locomotion agents [63]. Although these approaches showed improved performance compared to conventional DRL training, the policy architecture did not preserve the structure of the dynamical system defined by the CPG models. The work in [63] was extended in [64] to use recurrent neural networks. Authors in [64] fine-tuned locomotion priors using an evolutionary approach. Authors in [41], [65] treated CPG modules as part of the environment making it a black-box while training the policy, and pre-tuned their CPG modules either manually or by genetic algorithm. Although the work in [66] presented a similar approach, the policy architecture requires this method to execute the actions every time-step of the control task, and this study was also limited to a single-legged system for a hopping task.

In this work, we propose to address the limitations of previous works by developing a hierarchical locomotion policy architecture that embeds a CPG model (refer Fig. 2). We embed the locomotion policies, that we call DeepCPG, with Kuramoto oscillators [47] representing the CPG models. In this hierarchical setting, the artificial neural network predicts the parameters that define the CPG specifications, while the recurrent CPG layer outputs the action commands to the robots. We show the effectiveness of the proposed approach for developing the end-to-end control strategies for walking robots in high-dimensional observation and action spaces.

The rest of this paper is outlined as follows: Section II briefly discusses the RL basics. Section III provides the details of the proposed hierarchical policy architecture followed by Section IV describing the scaling strategy proposed for DeepCPG. Section V describes the training and deployment details for the proposed policy. Section VI presents the results and discussions. The conclusions and future directions are discussed in Section VII.

## II. BACKGROUND

The standard continuous control RL setting is adopted for the work reported in this paper and the terminology is adopted from the standard RL textbook by Sutton et al. [55]. The RL agent interacts with an environment according to a behavior policy $\pi$. The environment produces state (or observation) $s_t$ for each step $t$. The agent samples the action $u_t \sim \pi$ and applies it to the environment. For every action, the environment yields a reward $r_t$. The aim of the RL agent is given by Eq. (1):

$$\pi^* = \arg\max_{\pi} \mathbb{E}_{\tau \sim \pi, p_{s_0}} \left[ R(\tau) \right]$$
$$= \arg\max_{\pi} \mathbb{E}_{\tau \sim \pi, p_{s_0}} \left[ \sum_{t=0}^{\infty} \gamma^t r_t \mid \pi \right] \qquad (1)$$

where $\tau = (s_0, u_0, s_1, u_1, \dots)$ is the state-action trajectory sampled using policy $\pi$, the initial state is sampled from a fixed distribution $s_0 \sim p_{s_0}$ and $0 \leq \gamma < 1$ is the discount factor. $R(\tau)$ is the return of the agent over the complete episode.

One may efficiently learn a good policy from state-action-reward transitions collected by the RL agent by interaction with the environment. Temporal difference learning, a model-free RL approach, provides the framework to learn a control policy based on these collected interactions and by bootstrapping from the current estimation of the value function [55].

In our approach, we make use of the popular Twin Delayed Deep Deterministic Policy Gradient (TD3) algorithm [67] for learning in a continuous action space. This algorithm is a variant of Deep Deterministic Policy Gradient (DDPG) [68]. Using DDPG, a policy $\pi_\theta$ (parameterized by $\theta$) and the state-action value function $Q_\kappa$ (parameterized by $\kappa$) are concurrently learned. The Q-function represents the value of being in a state $s_t$ and taking action $u_t$, and it is trained to minimize the Bellman error over all the sampled transitions from the collected data given by Eq. (2). The policy $\pi_\theta$ is trained to maximize the $Q_\kappa(s_t, \pi_\theta(s_t))$ over all the observed states collected by environment interactions.

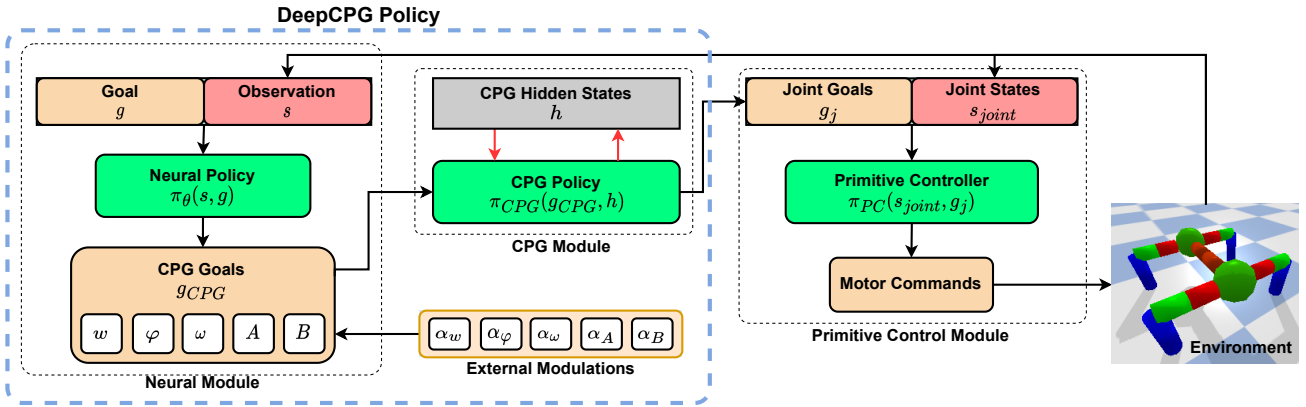

Fig. 2. Overview of Hierarchical Policy architecture used for Quadruped Robot

$$L(s_t, u_t, s_{t+1}) = (Q_\kappa(s_t, u_t) - r_t - \gamma Q_\kappa(s_{t+1}, \pi_\theta(s_{t+1})))^2 \quad (2)$$

Although TD3 and DDPG train a deterministic policy, experience replay is collected by augmenting the actions with Gaussian Noise. We note that TD3 makes several modifications to the DDPG algorithm to yield a robust and stable policy learning procedure. These modifications include (1) ensemble learning over Q-functions (two Q-functions are used), (2) policy and target networks update less frequently than Q-functions and (3) addition of noise to target actions, to make it harder for the control policy to exploit Q-function errors by smoothing out Q-values along with changes in action. Out of the two Q-functions $Q_{\kappa_1}$ and $Q_{\kappa_2}$ used for training the control policy in TD3, minimum of the two is chosen as a target value to prevent the over estimation [69]. For further details regarding TD3, we refer the readers to [67].

## III. APPROACH

In this work, we propose a hierarchical policy architecture to connect the DRL-based policies to CPG models. Figure 2 provides an overview of the control architecture using this policy. In this three-level hierarchy, the high-level policy is a neural network $\pi_\theta$ parameterized by $\theta$. The policy $\pi_\theta$ generates goals for the mid-level CPG policy $\pi_{CPG}$ that is based on the Kuramoto oscillator model [15]. The Kuramoto Model was a design choice but any other CPG models can be chosen without loss of generality. Finally, $\pi_{CPG}$ generates goals for the low-level primitive controller $\pi_{PC}$ that generates motor commands for the robot motor joints. The proposed method enables two objectives: (1) The neural network enables acting in the CPG parameter space, so the trajectories generated by CPGs remain smooth and can direct lower-level controllers generating joint motor commands for the robot safely as illustrated in Fig.1. (2) The CPG models allow embedding **intrinsic primitive behaviors** in the policy, enabling faster learning of the goal-directed behaviors.

### A. Neural Policy $\pi_\theta$

There are complex interactions between the central nervous system and the peripheral nervous system in animals. One of the basic and most important functions of the central nervous system is processing sensory feedback. Observations sensed from the environment need to be processed along with the agent's desired goals to generate appropriate responses. For the presented approach, this process is simulated using the neural policy $\pi_\theta$. This policy takes in the sensor feedback from the robot and the desired goal to generate the response $g_{CPG}$ that is relayed to the CPG policy $\pi_{CPG}$, where $g_{CPG}$ represents the parameters that drive the CPG behavior (see Fig. 2). Section III-B provides further details regarding this part of the system.

### B. Central Pattern Generators for Motor Trajectories

We simulate the central pattern generator for motor dynamics using the Kuramoto Model. Equations (3)-(6) describe this dynamical system. Although simple, this CPG network model enables replication of many behaviors observed in vertebrates. It also allows modeling of inter-oscillator couplings and assumes the presence of underlying oscillatory mechanisms. Simple modulation of CPG parameters can lead to emergence of various useful behaviors as shown in several studies [15]–[17], [70]. In this work, a single CPG controls a single joint on the robot.

$$\dot{\phi}_i = \alpha_\omega \omega_i + \sum_{i' \neq i} a_{i'} \alpha_w w_{ii'} \sin(\phi_{i'} - \phi_i - \alpha_\varphi \varphi_{ii'}) \quad (3)$$

$$\ddot{a}_i = \alpha_a(\beta_a(\alpha_A A_i - a_i) - \dot{a}_i) \quad (4)$$

$$\ddot{b}_i = \alpha_b(\beta_b(\alpha_B B_i - b_i) - \dot{b}_i) \quad (5)$$

$$y_i = b_i + a_i \sin(\phi_i) \quad (6)$$

In Equations (3)-(6), for CPG $i$, $\phi_i$ is the phase, $\omega_i$ is the natural frequency, $A_i$ is the desired amplitude and $B_i$ is the desired offset, $a_i$ is the current amplitude and $b_i$ is the current offset. $w_{ii'} \geq 0$ represents the coupling weight between the incoming connection from CPG $i'$ to CPG $i$, and $\varphi_{ii'}$ is the connection phase bias between the CPGs $i$ and $i'$. Parameters $(\alpha_a, \beta_a)$ and $(\alpha_b, \beta_b)$ are constants, whereas $\dot{\phi}_i, \dot{a}_i, \dot{b}_i$ represents the first time derivatives and $\ddot{a}_i, \ddot{b}_i$ the second time derivatives of respective variables. The external modulation constants $\alpha_x \geq 0 \forall x \in \{w, \varphi, \omega, A, B\}$ account for possibly user-defined external influence on the parameters that drive the $\pi_{CPG}$ policy. For instance, $\alpha_A$ and $\alpha_B$ could depend on the motor joint limits on the robot and could be set to match these limits. Parameters $\alpha_w, \alpha_\varphi$ and $\alpha_\omega$ influence the rate of change of joint angles. We kept these values constant for the robot in our implementation. The $\pi_\theta$ generates $g_{CPG} \equiv \{w, \varphi, \omega, A, B\}$ to govern the CPG behavior for the robot. Thus, $A_i \in A$ is then scaled by $\alpha_A$ depending on the system requirement after a prediction from $\pi_\theta$. The output of the CPG $i$ is given by $y_i$.

The CPG network produces the desired joint trajectories for all the motor joints on the robot. The parameters $(w_{ii'}, \varphi_{ii'}, \omega_i, A_i, B_i)$ define the behavior for CPG node $i$ in this network. We assume that each node in the CPG network influences every other node symmetrically. Therefore, for a robot with $N$ motor joints, we have a symmetric weight matrix $w = \{w_{ii'}\} \forall i, i' \in \{1, \ldots, N\}$ is $N \times N$ with zero diagonal elements, a phase bias matrix $\varphi = \{\varphi_{ii'}\} \forall i, i' \in \{1, \ldots, N\}$ that is $N \times N$ skew-symmetric, and natural frequencies $\omega$, desired amplitudes $A$, and desired offsets $B$ each forming $N$-dimensional vectors. The zero diagonals of matrices $\varphi$ and $w$ signify that CPG nodes do not have recurrent connections to themselves and the network formed by these nodes is a bidirectional network. In this bidirectional CPG network, each CPG node $i$ influences its neighboring CPG node $i'$ proportional to the weight $w_{ii'}$. Node $i$ being out of phase to node $i'$ by $\varphi_{ii'}$, makes node $i'$ out of phase to node $i$ by $-\varphi_{ii'}$ as a result of skew-symmetric matrix $\varphi$. Additionally, the assumption of a symmetric $w$ and skew-symmetric $\varphi$ halves the total number of neural outputs predicted by $\pi_\theta$.

In Fig. 2, the set of parameters $\{w, \varphi, \omega, A, B\}$ are also referred to as CPG Goals $g_{CPG}$. The value of $g_{CPG}$ is predicted using neural policy $\pi_\theta$. Together, $g_{CPG}$ and the Equations (3)-(6) form the CPG Policy $\pi_{CPG}$. This $\pi_{CPG}$ consists of time dependent differential equations and maintains its hidden state $h_t$ for each step $t$ that consists of $\{\phi_t, \dot{\phi}_t, a_t, \dot{a}_t, \ddot{a}_t, b_t, \dot{b}_t, \ddot{b}_t\}$. The CPG network outputs the desired motor joint states $g_j = y(t)$ (referred as Joint Goals in Fig. 2) that is a vector of $N$ dimensions for a robot with $N$ motor joints at each step $t$. To reduce notation, we refer to the vector formed by concatenation of goal and robot observations $(g, o)$ as $o$ unless otherwise stated. It should be noted that $g_{CPG}$ governs the entire architecture of CPG network formed by $\pi_{CPG}$ based on the observed robot state.

### C. Primitive Controller $\pi_{PC}$

As Fig. 2 shows, the outputs action $g_j$ of $\pi_{CPG}$ form the desired states of the robot joints. The value of $g_j$ is relayed to the lower level primitive controller $\pi_{PC}$ that generates the motor commands. Without loss of generality, $\pi_{PC}$ can be any type of controller such as a Proportional-Integral-Derivative (PID) controller or another neural policy. Policy $\pi_{PC}$ can be designed as an inverse controller similar to the one described in [60]. To keep our implementation simple, we chose the PD-controller as $\pi_{PC}$. The proportional $k_p$ and derivative $k_d$ gains of the PD controller were manually tuned and kept equal for all the robot joints in all the experiments.

## IV. SCALING DEEPCPG POLICIES

To enable scalability of DeepCPG policies, we propose to train these policies using a multi-agent RL setup [71]. In this setup, we define the modular robot control policy as a set of modular agents that *cooperatively* solve the desired task. A similar strategy for training multi-legged robot locomotion policy has been proposed in [18], [72]. The primary difference between these works and our approach lies in the trained policy where we learn the neural policy to predict CPG model parameters. Therefore, a fully cooperative multi-agent task is formulated such that a team of modular agents forming the control policy interact with the same environment (modular robot) to achieve a common goal. For modular robot configuration, we consider $n$ agents corresponding to $n$ modules in the system. Figure 10 visually illustrates a single robot divided in multiple modules. In this Markov game, $\mathbf{s} \in \mathbf{S}$ describes the true state of the modular robot. Each module $m \in \{1, \ldots, n\}$ consists of its corresponding action set $U_m$ and (partial) observation set $O_m$. A module $m$ uses a policy $\pi_{\theta_m}$ (stochastic or deterministic) to select a continuous (or discrete) action $u_m \in U_m$ at each step. This produces a next state $\mathbf{s}'$ according to state transition function $\mathbf{P}(\mathbf{s}'|\mathbf{s}, \mathbf{u}) : \mathbf{S} \times \mathbf{U} \times \mathbf{S} \mapsto [0, 1]$, where joint action $\mathbf{u} \in \mathbf{U} \equiv \{U_m\} \forall m \in \{1, \ldots, n\}$. Modular agents earn a joint reward $r(\mathbf{s}, \mathbf{u})$ for every action $\mathbf{u} \equiv \{u_1, \ldots, u_n\}$ taken in state $\mathbf{s}$. Thus, this Markov Decision Process (MDP) can be summarized in a tuple $\langle n, \mathbf{S}, \mathbf{U}, \mathbf{P}, r, \mathbf{O}, \gamma, \mathbf{p_{s_0}} \rangle$. The set of observations corresponding to each module is contained in $\mathbf{O} \equiv \{O_m\} \forall m \in \{1, \ldots, n\}$. Each modular agent in the robot learns policies $\pi_{\theta_m}$ conditioned on its local observation correlated with the true state $o_m = \mathbf{O}(\mathbf{s}, m)$ where $o_m \in O_m$ and $s \in \mathbf{S}$ (we will use $\pi_m$ instead of $\pi_{\theta_m}$ to avoid notation clutter). The distribution of initial states of the system is $\mathbf{p_{s_0}} : \mathbf{S} \mapsto [0, 1]$. The overall objective is to maximize the sum of discounted return $R_t = \sum_{t'=0}^{\infty} \gamma^{t'} r_{t+t'}$ where $\gamma$ is the discount factor.

Figure 3 provides a schematic overview of this setup for a modular robot with $n$ modules. Each module $m \in \{1, \ldots, n\}$ consists of its corresponding action set $U_m$ and (partial) observation set $O_m$. The module actor $\pi_{\theta_m}$ corresponds to the $m$-th robot module. The output of $\pi_{\theta_m}$ is sent to $\pi_{CPG,m}$ that generates action commands $u_m = g_{j,m}$ for the Module-$m$ of the robot. To enable emergence of coordination in the modular robot, we define the observation vector perceived by each module $m$ with three distinct components. These three components comprise of global contextual information $o_g$, local contextual information $o_{(m)}$ private to module $m$, and inter-modular contextual information $\{o_{(m,m')}\} \forall m' \in \{1, \ldots, n\}$.

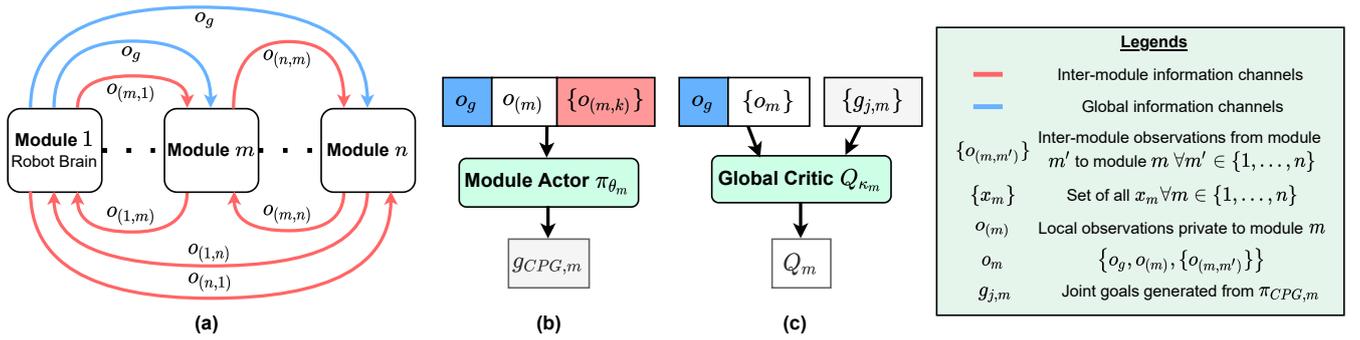

Fig. 3. (a) Schematic of information sharing across various modules in the modular robot; (b) Actor or Policy for $m$-th module parameterized by $\theta_m$; (c) Centralized critic or action value function for $m$-th module parameterized by $\kappa_m$

The global information $o_g$ consists of information about the desired goal of the system.

To train this modular system, a centralized training with decentralized execution setup is adopted [73], [74]. The independent training of modular policies may lead to poor performance [75]. The major issue leading to such performance degradation is the non-stationarity of environments that prevents the use of the trajectory history of the agents during training. Thus, in the regime of centralized training, policy learning can be stabilized by the use of extra global information during training, but during execution, each modular agent uses only its local action-observation history. For centralized training, the joint policy, denoted by $\Pi$, induces a joint action-value function $Q_{\kappa_m}^{\Pi}(\mathbf{s}, \mathbf{u}) = \mathbb{E}_{\mathbf{s}'}\left[r(\mathbf{s}, \mathbf{u}) + \gamma \mathbb{E}_{\mathbf{u}' \sim \Pi}[Q_{\kappa_m}^{\Pi}(\mathbf{s}', \mathbf{u}')]\right]$ (henceforth, we will replace $Q_{\kappa_m}$ with $Q_m$ to avoid notation clutter). It should be noted that this joint policy is a set $\Pi \equiv \{\pi_1, \ldots, \pi_n\}$. The primary advantage of this action-value function is that, if we know the actions taken by all modules along with their true states during training, the environment is stationary even as the individual policies $\pi_m$ are being updated [74]. The joint action value function $Q_m^{\Pi}$ is trained by minimizing the a mean-squared Bellman loss function, which estimates how close $Q_m^{\Pi}$ comes to satisfying the Bellman equation. This loss function is given by Eq.(7):

$$\mathcal{L}(\kappa_m) = \mathbb{E}_{\mathcal{R}}\left[\left(Q_m^{\Pi}(\mathbf{s}, \mathbf{u}) - (r_m(\mathbf{s}, \mathbf{u}) + \gamma Q_m^{\Pi^{targ}}(\mathbf{s}', \mathbf{u}'))\right)^2\right] \quad (7)$$

where $r_m(\mathbf{s}, \mathbf{u})$ is the reward received by module $m$. In this case, the collaborative task of all the modules corresponds to a unified reward $r(\mathbf{s}, \mathbf{u})$ measuring the performance of the robot as a whole. Thus, it is not necessary to consider a different reward function for each modular agent. The set of target policies $\Pi^{targ}$ with delayed parameters $\{\theta_1^{targ}, \ldots, \theta_n^{targ}\}$ produces $\mathbf{u}' \equiv \{u_1', \ldots, u_n'\}$. $Q_m^{\Pi^{targ}}$ corresponds to the target critic with delayed parameters $\kappa_m^{targ}$. $\mathcal{R}$ is the replay buffer containing the transition tuples $(\mathbf{s}, \mathbf{u}, \mathbf{s}', r)$. We can evaluate the policy gradient for each module $m$ according to Eq. (8):

$$\nabla_{\theta_m} J(\pi_m) = \mathbb{E}_{\mathcal{R}}\left[\nabla_{\theta_m} \pi_m(o_m) \nabla_{u_m} Q_m^{\Pi}(\mathbf{s}, \mathbf{u})\right] \quad (8)$$

where $Q_m^{\Pi}(\mathbf{s}, \mathbf{u})$ is a centralized action value function. Action $u_m$ is obtained from its current policy $\pi_m$ and remaining $\{u_{m'}\} \forall m' \neq m$ are obtained from the replay buffer $\mathcal{R}$.

Readers should note that we provide the explanation for $n$ modules in a modular system for the sake of completeness. In the experimental section, we restrict ourselves to 2 agents (refer to Fig. 10) to demonstrate the proof-of-concept using the DeepCPG policies in this proposed multi-agent RL framework. Additionally, for locomotion tasks, the velocity direction in the body reference frame of Module-1 was used as $o_g$ in observation. The local modular observation of the two modules considered in the modular robot in Fig. 10 is given by sets $o_1 = \left\{o_g, o_{(1)}, \{o_{(1,1)}, o_{(1,2)}\}\right\}$ and $o_2 = \left\{o_g, o_{(2)}, \{o_{(2,1)}, o_{(2,2)}\}\right\}$. We keep $o_{(1,1)} = o_{(2,1)}$ and $o_{(1,2)} = o_{(2,2)}$.

## V. TRAINING DEEPCPG POLICIES

As described in Section II, we use the TD3 algorithm for training the control policy. In the DeepCPG policies using this approach, the policy gradient calculated in TD3 must back-propagate through actions $u = g_j$ generated by $\pi_{CPG}$ followed by $\pi_\theta$. To back propagate, $\pi_{CPG}$ must be differentiable in practice. While the Kuramoto model based CPG network is differentiable, to find the analytical solution of the set of differential equations 3-6, the integration is implemented with discrete steps of length $\delta t$. For backpropagation to work, it can be shown that, the derivatives of these equations with respect to the parameters predicted by $\pi_\theta$ in Eq. (9) exist:

$$\frac{\partial y_i}{\partial w_{ii'}}, \quad \frac{\partial y_i}{\partial \varphi_{ii'}}, \quad \frac{\partial y_i}{\partial \omega_i}, \quad \frac{\partial y_i}{\partial A_i}, \quad \frac{\partial y_i}{\partial B_i} \quad (9)$$

The complete derivation for Eq. (9) is provided in Appendix.

Given that the derivatives of the $\pi_{CPG}$ outputs with respect to the parameters $g_{CPG} \sim \pi_\theta$ exist, it is possible to train a DRL policy in end-to-end fashion to predict CPG actions given the observations from the environment. As discussed in Section II, we use the TD3 algorithm to train $\pi_\theta$ by propagating the policy gradients through $\pi_{CPG}$. However, a different RL algorithm for training the policy can also be used.

As shown in Fig. 4, policy $\pi_\theta$ is modeled with multiple heads each corresponding to the CPG parameters $g_{CPG} \subset \{w_{ij}, \varphi_{ii'}, \omega_i, A_i, B_i\}$. It should be noted that it is possible to train $\pi_\theta$ to predict (1) all the CPG parameters $g_{CPG}$ or (2) only a subset of $g_{CPG}$ with the remaining parameters preset manually to specific values.

In this actor-critic method, two state-action value functions $Q_{\kappa_1}$ parameterized by $\kappa_1$ and $Q_{\kappa_2}$ parameterized by $\kappa_2$ are learned as critics. The job of these critics is to criticize the actions taken by the policy in the given states. The critique or the values given by the critics takes the form of a Temporal Difference error (see Eq. 2). The output signals of the critics drive all learning in both actor (policy) and critic. For critic architectures, it is possible to estimate the state-action values using the actions $g_{CPG}$ by higher-level policy $\pi_\theta$ or actions $g_j$ generated by the CPG network $\pi_{CPG}$. Based on our experiments, we observed that a better policy is trained when we use $g_j$ for action-value prediction. This may be because the behavior of $\pi_{CPG}$ is modified for each step of higher-level policy $\pi_\theta$. This changing behavior of the lower-level policy $\pi_{CPG}$ creates a *non-stationary* environment for the higher-level policy $\pi_\theta$, and old off-policy experiences may exhibit different transitions conditioned on the same goals. However, critiquing the actions $g_j$ may alleviate this problem as critics are able to observe the complete behavior of the actor that consists of $\pi_\theta$ and $\pi_{CPG}$.

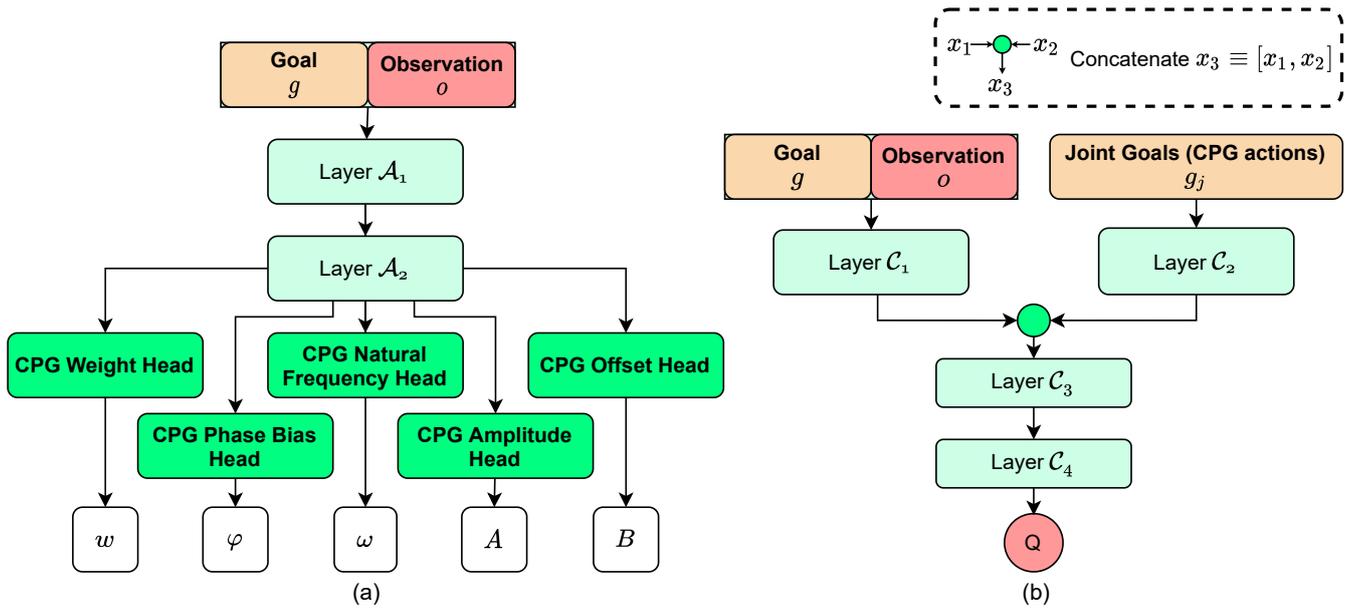

Fig. 4. Schematic of Neural Network architectures: (a) Neural Policy $\pi_\theta$; (b) Critics $Q_{\kappa_1}$ and $Q_{\kappa_2}$. In the article, the observation vector $(g, o)$ is represented as $o$ to reduce notation unless otherwise stated. Note: The *Heads* in $\pi_\theta$ are the layers in the neural network.

As discussed in Section III-B, the weight matrix $w$ is constrained to be symmetric with a zero diagonal. and the phase bias matrix $\varphi$ is constrained to be skew symmetric for the CPG network $\pi_{CPG}$. Thus, to maintain these constraints, the actor $\pi_\theta$ is designed to predict only the off-diagonal (upper triangular) elements with an offset of 1, i.e., the diagonal elements are not included in the prediction vector. Thus, for robot with $N$ CPG nodes in $\pi_{CPG}$, the weight and phase bias predictions are $\frac{N(N-1)}{2}$ dimensional in size. These prediction vectors are then converted to corresponding $w$ and $\varphi$ matrices of shape $N \times N$ inside $\pi_{CPG}$. To keep the outputs of $\pi_\theta$ bounded, we use the 'tanh' activation function for all the output heads.

$$y = 0.5\big(x(y_{max} - y_{min}) + (y_{max} + y_{min})\big) \quad (10)$$

The affine transformation shown in Eq. 10 is applied to predictions of $\pi_\theta$ to transform them within the respective bounds, i.e., $0 \le w_{ii'} \le 1, -1 \le \varphi_{ii'} \le 1, 0 \le \omega_i \le 1, 0 \le A_i \le 1, -1 \le B_i \le 1$ before passing in $\pi_{CPG}$.

We provide the complete pseudo-code for training the DeepCPG policy using DRL in Appendix. The input to policy $\pi_\theta$ is a sequence of states of length $\tau_o$, i.e., in Fig. 2 and Fig. 4, $s = s_{t-\tau_o:t}$. The CPG policy acts on the environment for $\tau_c$ steps for each step of $\pi_\theta$.

*Scaling DeepCPG policies*

Based on the setup described in Section IV, we apply a multi-agent RL algorithm to learn a scalable control policy for a modular robot. To that end, we use the DDPG algorithm customized for a multi-agent setup [74]. In the multi-agent setup, we chose DDPG over TD3 for training to reduce the compute requirement. We train the modular robot by bootstrapping the policies learned during a simpler design stage of the system, and use it on the modular system with added design complexity as a result of Module-2 attached to Module-1. Readers should note that the complexity in these modular systems is defined in terms of their degrees of freedom. The system with a higher count of motor joints is considered a comparatively more complex system than a system with a lower number of motor joints. An explicit increase in complexity in terms of motor joints also results in an increase of phenotypic complexity, where the robot body itself changes as a result of additional module. The scaling of policy and the critic in terms of this modular setup is given in Fig. 5.

*Deployment*

The proposed method uses trained policy $\pi_\theta$ once per $\tau_c$ steps in robot environment. This achieves speedup during inference as fewer forward passes are required for a DeepCPG policy

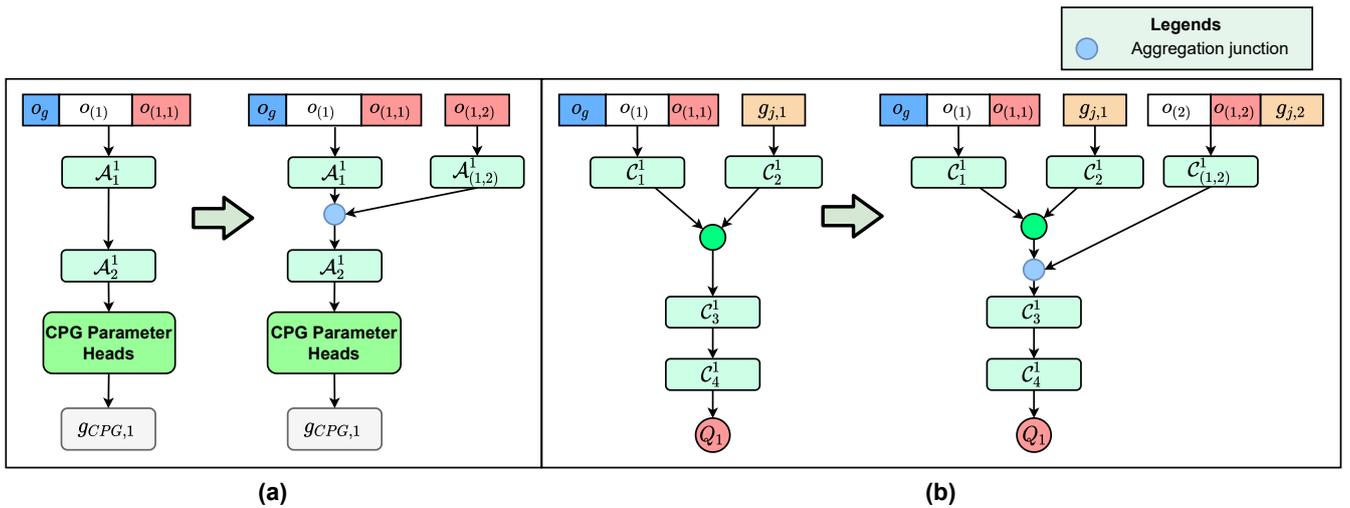

**(a)** **(b)**

Fig. 5. Neural Network scaling: (a) Scaling of $\pi_{\theta_1}$ when Robot Module-2 is attached to Module-1; (b) Scaling of centralized critic of Module-1 when Robot Module-2 is attached to Module-1. The same architecture is also used for Module-2 policy and centralized critic. Note: $\mathcal{A}_l^m$ is a layer $l$ of policy corresponding to module $m$ and $\mathcal{C}_l^m$ is layer $l$ of critic corresponding to module $m$.

as compared to a feed-forward multi-layered perceptron based policy. Given the limited computational power of the real robot, this skip of $\tau_c$-steps also facilitates real-time processing. The pseudo-code for the policy deployment is provided in Appendix. The two critics learned during the training process are not needed during the deployment for inferring from the policy, further lowering the computational cost.

## VI. RESULTS AND DISCUSSION

Experiments were carried out using a physics-based simulation environment as well as physical real-world robots. This section presents the experimental setup and results for the evaluation of DeepCPG policies. We evaluate the CPG policies on intrinsically motivated locomotion task (VI-B), Go-to-Goal task (VI-C) and visual feedback task (VI-F).

### A. Experimental Setup - Physics Based Engine

For developing a modular system using legged robots, we chose the Bullet Physics engine [76]. The quadruped robot simulation was developed to perform experiments with the proposed algorithm. Each joint in this robot is a revolute joint. Figure 6 shows an image and the corresponding schematic of the quadruped robot used in the evaluation of the proposed work. This robot is a 12-degree-of-freedom system (12 active joints) as indicated in this figure.

The simulated environments used in the experiments were wrapped with the OpenAI gym interface [77]. Different features of the environments developed for testing the proposed approach are also shown in Fig. 6. PyTorch was used as the deep learning framework [78]. The simulations were performed on a machine with an Intel-i7 processor, 16GB RAM, and NVIDIA RTX 2070. The training of each experiment took approximately 6 hours for observations without images. For experiments that used visual feedback for the robot, each run took approximately 23 hours.

To evaluate the proposed approach, we provide the comparison with a baseline of feed-forward policy trained using the TD3-algorithm [67]. The base architectures of both, feed-forward policy and DeepCPG policy, were kept identical (Note: base architecture refers to the network architecture before the output layer). For the feed-forward policy, the output layer consisted of a fully connected layer of neurons as opposed to the one in the DeepCPG policy that consisted of the CPG model. The dimensions of this layer were kept equal to the action-space dimensions of the robot. Further details about the hyperparameters used for training the RL policies are given in Table I. Each experiment was performed five times with different random seeds for a random number generator. The resultant statistics are provided in the plots.

Figure 4 provides the architectural details of the neural networks corresponding to $\pi_\theta$. For the feed-forward neural network, layer $\mathcal{A}_1$ consisted of 1024 neurons with ReLU activation and $\mathcal{A}_2$ consisted of 512 neurons with ReLU activation. CPG parameter heads for $w, \varphi, \omega, A, B$ are layers with 512 neurons and ReLU activation. The input dimensions of the network depend on the observation space dimensions of the robot and the output dimensions depend on the architecture of $\pi_{CPG}$ and the action space of the robot as discussed in Section V. For the Critic architecture, $\mathcal{C}_1$ and $\mathcal{C}_2$ each consisted of 1024 neurons with ReLU activation, $\mathcal{C}_3$ consisted of 512 neurons with ReLU activation, followed by a linear layer $\mathcal{C}_4$ with 512 neurons. The critics output a scalar value. It should be noted that for training the actor with visual feedback (images as observations), actor and critic architectures were modified to include 2 convolutional layers before $\mathcal{A}_1$ and $\mathcal{C}_1$, respectively. The image size used during training was $32 \times 32$. These images were converted to grayscale before feeding them to the neural networks. The images for $\tau_o$ steps were stacked and passed through the two convolutional layers and then flattened and concatenated with other observations (if any) to pass through the first linear layer of 1024 neurons, shown in Fig. 4, for the actor and both the critics. The first 2D-convolutional layer consisted of $\tau_o$ input channels and 10 output channels. For the second 2D-convolutional layer, 10 input channels, 15 output channels were used. The kernel size of 3 and stride of 1 were used for both the layers.

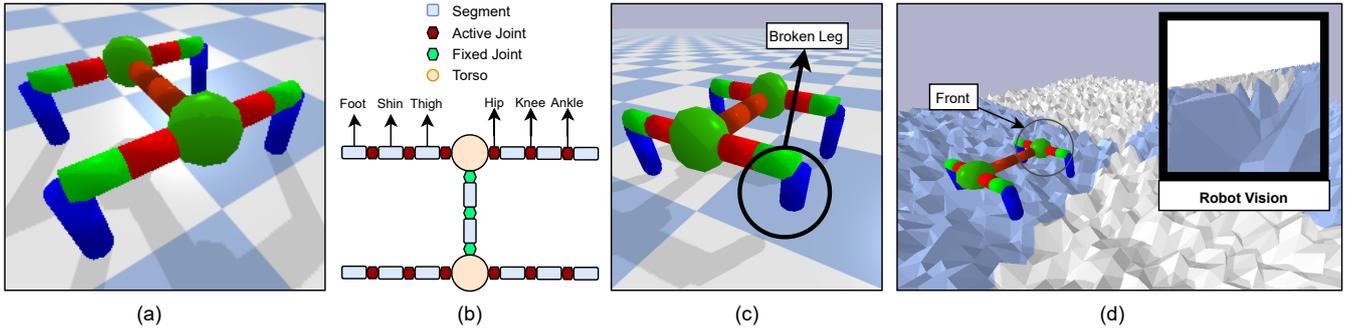

Fig. 6. Quadruped robot: (a) Bullet Physics GUI; (b) Schematic; (c) Fault in front-left limb; (d) Uneven terrain and with simulated visual feedback.

TABLE I
Hyperparameters for training DeepCPG policies

| Parameter | Value |
|---|---|
| Learning rate $\eta$ for actor and critics | $2 \times 10^{-4}$ |
| Neural network optimizer | Adam [79] |
| Adam parameters $(\beta_1, \beta_2)$ | $(0.9, 0.999)$ |
| Discount factor $\gamma$ | 0.95 |
| Max. episode length $T$ | 2000 |
| Batch size $|B|$ | 64 |
| Replay buffer size $|\mathcal{R}|$ | $1 \times 10^6$ |
| Gradient norm clipping threshold | 2.0 |
| Babbling steps $\tau_b$ | 10000 |
| Update steps $\tau_{update}$ | 1 |
| Policy update delay $\tau_{delay}$ | 2 |
| CPG Policy steps $\tau_c$ | 5 |
| Observation steps $\tau_o$ | 5 |
| $\{\alpha_w, \alpha_\varphi, \alpha_\omega, \alpha_A, \alpha_B\}$ | $\{600, \pi, 20, 0.8, 0.2\}$ |

### B. Intrinsic Motivation Task

Good exploration enables the policy to discover various behaviors. In practice, to enable this, the learning agent should leverage information gained from environment dynamics. To encourage this outcome, we define an intrinsic reward function in equation (11) that encourages the policy to learn the locomotion behavior. The agent is rewarded for successfully moving the robot in $X - Y$ plane.

$$r_t = c_v ||v_{(x,y),t}||_2 - c_\vartheta ||\vartheta_t||_2 - c_z ||z_t||_2 - c_j ||s_{joint_t}||_2 + c_b \quad (11)$$

In Eq.11, $v_{(x,y),t}$ denotes the current velocity of the robot along $x$ and $y$ direction. $\vartheta_t$ is the angular velocity of the robot. $s_{joint_t}$ represents the vector of joint angles of the robot at time $t$. $z_t$ represents the robot height above the ground at time $t$. The term with $s_{joint_t}$ implicitly regularizes actions by imposing a penalty on the robot joint angles. The last term $c_b$ contributes to the bonus value for the robot for staying alive. The coefficients $c_{(\cdot)} > 0 \forall \{v, \vartheta, z, b, j\}$. It should be noted that Eq.11 does not contribute to a goal directed behavior of any kind in the learned policy. In the experiments, $c_v = 2.0$, $c_b = 4.0$, $c_\vartheta = 0.5$, $c_z = 5.0$, $c_j = 10^{-3}$. Figure 7-(a) provides the plot of episode returns over the training of the robot. From this plot, it can be observed that the DeepCPG policy performance is on par with the feedforward policy trained using TD3. The advantage of the CPG priors in the locomotion is evident from the results in Section VI-C, where these policies are fine-tuned for the goal-directed task.

### C. Go-to-Goal Task

The policies that were trained in Section VI-B were fine-tuned to learn a goal-reaching behavior with the modification to the reward in Eq. (11). The weights that were learned for the policies for the task in Section VI-B were transferred for this downstream task. Equation (12) provides the updated reward function used for learning a go-to-goal behavior.

$$r_t = c_e(v_{(x,y),t} \cdot \hat{e}_g) + c_v ||v_{(x,y),t}||_2 - c_\vartheta ||\vartheta_t||_2 \\ - c_z ||z_t||_2 - c_j ||s_{joint_t}||_2 + c_b \quad (12)$$

In Equation (12), $\hat{e}_g$ is the unit vector pointing towards goal $(x_g, y_g)$ from the robot position $(x_t, y_t)$ and $c_{(\cdot)} > 0 \forall \{e, v, \vartheta, z, j, b\}$. For the experiments, $c_e = 5.0$. The values of the remaining coefficients are provided in Section VI-B.

The advantage of the DeepCPG policy is visible from the training plots in Fig. 7-(b) in terms of sample efficiency. It was able to learn the go-to-goal behavior faster as compared to the baseline policy. This could be attributed to the behavioral priors in the DeepCPG policy as a result of $\pi_{CPG}$.

### D. Ablation Study: Fault Tolerance

To investigate the fault-tolerant behavior of the proposed policy, we introduced a fault in the robot. This environment change is shown in Fig. 6-(c). One of the legs in the robot was broken. The reward function was updated from Eq. (11) to Eq. (13). In Eq. (13), the robot was evaluated based on its performance to walk along the x-axis, so its velocity along x-axis $v_{(x),t}$ was rewarded. All the coefficient values were kept as defined in Section VI-B.

$$r_t = c_v ||v_{(x),t}||_2 - c_\vartheta ||\vartheta_t||_2 - c_z ||z_t||_2 - c_j ||s_{joint_t}||_2 + c_b \quad (13)$$

Figure 7-(c) shows the episode return plots for the policies from Section VI-B fine-tuned for the system with the broken-leg. The DeepCPG policy shows comparatively better performance than the feed-forward policy. As compared to the system without any

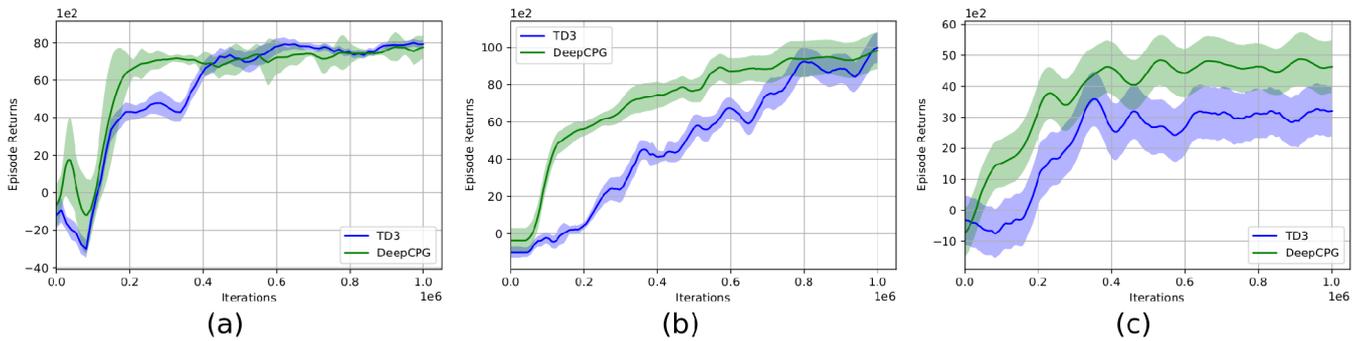

Fig. 7. Plot of episode returns the task of: (a) moving as fast as possible; (b) go-to-goal; (c) recovery from the fault. Legends: (TD3) Feed-forward policy trained using TD3, (DeepCPG) DeepCPG policy.

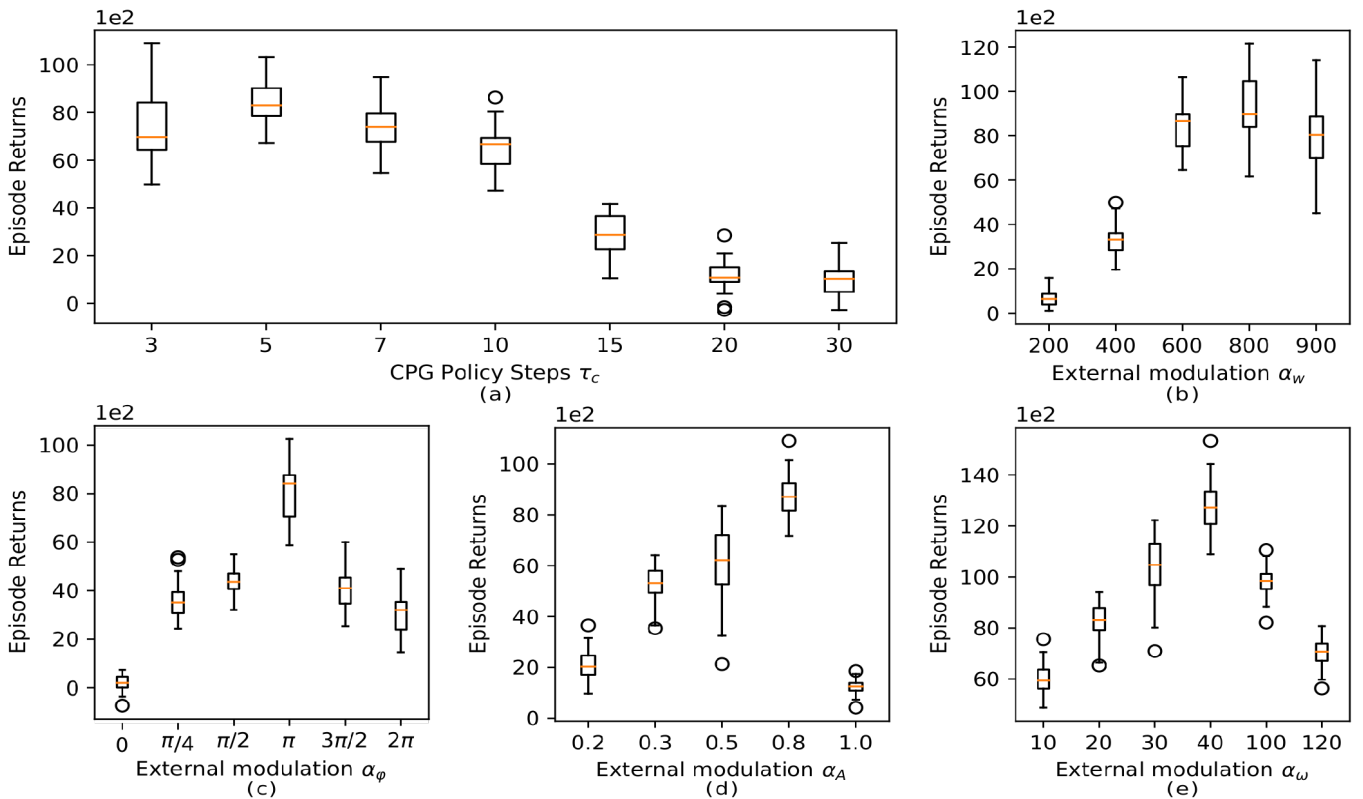

Fig. 8. Plot of episode returns with respect to respective parameter values used in the ablation study: (a) CPG Policy Steps $\tau_c$; (b) External modulation of CPGs using $\alpha_w$; (c) External modulation of CPGs using $\alpha_\varphi$; (d) External modulation of CPGs using $\alpha_A$; (e) External modulation of CPGs using $\alpha_\omega$

faults, the performance of this system saturated to comparatively lower episode returns as observed from the plots. Nonetheless, the DeepCPG policy converged to a comparatively higher performance than the feed-forward policy.

### E. Ablation Study: CPG Parameters

Figure 8 presents the ablation studies of various parameters used in the trained DeepCPG policy. The policy trained for the experiments described in Section VI-B is used for this study. The plots in Fig. 8 were generated by running each experiment for 30 episodes with random initialization of the robot. Only the parameter under consideration was varied in each case. The values of the remaining parameters were kept equal to those used during training, as given in Table I.

In Fig. 8-(a), we varied the parameter $\tau_c$ during the deployment of the policy. A higher value of this parameter signifies a lower execution frequency of $\pi_\theta$ in the DeepCPG policy. It can be observed from the plots that the trained policy was robust to variation $\tau_c$ during deployment for a wide range of values. Although $\tau_c = 5$ was used during training, similar performance was observed for $\tau_c = 10$. The performance of the robot degraded when the value of $\tau_c$ was further increased. As the robot's response to the sensor-feedback was delayed due to reduced execution frequency of $\pi_\theta$ when $\tau_c$ increases, we observed a corresponding performance degradation.

Figure 8-(b) presents the results of episode returns from the trained policy with variation of $\alpha_w$ in Eq. (3). The higher value of this parameter enables faster synchronization of all the oscillators in Kuramoto CPG nodes contained in $\pi_{CPG}$. The plot shows a corresponding response of the trained DeepCPG policy with the variation of this value. The policy was trained using $\alpha_w = 600$. As the CPG values are not able to synchronize when $\alpha < 600$, the policy performance was observed to degrade. For $\alpha > 600$, the trained DeepCPG policy showed robustness and its performance did not suffer.

Figure 8-(c) shows the results for variation of $\alpha_\varphi$ in Eq. (3). The trained policy performance was observed to be sensitive to variation of $\alpha_\varphi$. When $\alpha_\varphi = 0$, the phase bias of each CPG node of the $\pi_{CPG}$ becomes zero, the robot was not able to perform well.

Figure 8-(d) shows the results for variation of $\alpha_A$ in Eq. (4). It should be noted that we maintained a constraint of $\alpha_A + \alpha_B = 1$ to ensure $\max(\alpha_A A_i) + \max(\alpha_B B_i) \leq \max(s_{joint_i})$ where $\max(s_{joint_i})$ corresponds to the limits of $joint_i$ on the robot and $A_i$ and $B_i$, respectively, correspond to the desired amplitude and desired offset of the CPG node corresponding to that joint. This constrains the joint movement resulting from a CPG node to be within the joint's physical limits. From the plot shown in Fig. 8-(d), the robot achieved less reward for lower values of $\alpha_A$. This was the result of the smaller amplitudes of a robot's gait. On the other hand, an interesting observation is for $\alpha_A = 1$ where the robot performance degrades. We observed that the offset is an important component in DeepCPG Policy. When $\alpha_A = 1$, $\alpha_B = 0$ resulting in the performance degradation.

Figure 8-(e) shows the results for variation of $\alpha_\omega$ in Eq. (3). This parameter influences the oscillation frequency of the CPG nodes in DeepCPG policies. The policy showed robust behavior to some variation of $\alpha_\omega$. The higher the value of $\alpha_\omega$, the higher the oscillation frequency of CPG nodes. This increase in frequency is reflected in higher robot speeds and a resultant increase in episode return. At high values, $\alpha_\omega = 100$ and $\alpha_\omega = 120$, the policy performance was degraded. This may be associated with the non-stationarity in the observation space with high $\alpha_\omega$. Although the trained policy was robust to some variations of this parameter, the degrading performance could be the result of distribution drift in the observations space when $\alpha_\omega$ increases.

Overall, it can be observed from the parameter study conducted in this section that the DeepCPG policy is robust to a wide range of parameter changes during deployment even when these parameters are not shown to the policy during the training phase.

### F. Visual Feedback Task

We also evaluated the DeepCPG policy on the environment designed with high-dimensional image observations. Two cases were tested for DeepCPG and feed-forward policies each: (1) The robot was provided with visual feedback only, (2) The robot was provided with visual feedback along with input from the proprioceptive touch sensors on its feet. The sample image of the environment is shown in Fig. 6. The terrain was randomized in every episode. This randomization was introduced to enable the generalization of the quadruped to walk on rough terrain. The terrain height was randomly sampled from the range $[0, 0.1 \times h_r]$ where $h_r$ represents the robot height.

The robot was trained with the reward function defined in Eq. (13) with along the x-axis. Figure 9-(a) shows the results of the experiments with visual feedback to the robot. From these results, it can be observed that the learned DeepCPG policy was able achieve comparatively higher reward than the feed-forward policy.

In general, it was observed that the feed-forward actor showed comparatively poor performance than DeepCPG for any environment perturbation. The feed-forward actor consistently demonstrated lower asymptotic performance, especially in the case of visual feedback with no proprioceptive feet touch sensors (see Fig. 9-(a)).

### G. Energy Analysis

This section presents the results for the energy analysis of the walking robot described in the previous sections. We compare the results generated from the feed-forward MLP policy with the DeepCPG policy. Both the policies were trained with the reward function Eq. (11) in Section VI-B.

Figure 9-(b) provides the joint trajectories for the robot trained using these policies recorded for a span of 3 seconds arbitrarily from the simulation. It should be noted that these trajectories are plotted using the joint angles of the robot simulation and do not correspond to the motor commands sent from the neural network. From this figure, it can be observed that the DeepCPG policy enables the generation of smooth joint trajectories. This is attributed to the CPG dynamical system embedded as an output neural layer. This layer can predict in the trajectory space of the Kuramoto Model of CPGs and thus the joint trajectories generated from these commands appear smooth. On the other hand, for Feed-forward MLP, the neural network predicts the values of the motor commands directly for every iteration. As a result of this, the joint trajectories appear non-smooth. The average energy consumed by all the motor

TABLE II
Energy expended by the quadruped robot joints

| Policy | Value | $t$ | $T$ |
|--------|-------|-----|-----|
| FF-MLP | $486.78 \pm 68.92$ J | $43 \pm 13.24ms$ | $48 \pm 13s$ |
| DeepCPG | $\mathbf{253.75 \pm 20.16}$ J | $46 \pm 26.19ms$ | $41 \pm 22s$ |

joints on the robot in the simulation is shown in Table II where Column $t$ denotes time taken by the policies for each iteration (in milliseconds) and Column $T$ denotes the time taken for task completion (in seconds). We estimate these values by using the equation $Work = \sum_{joint} \sum_t Torque_{joint_t} \cdot \partial s_{joint_t}$, where $Torque_{joint_t}$ is the torque applied on a motor joint at time $t$ and $\partial s_{joint_t}$ is the change of joint state from time $t$ to $t+1$. This statistic was evaluated based on the data gathered over the complete episode of $2,500$ steps for 5 episodes. Given the non-smooth nature of the joint trajectories generated from Feed-Forward MLP policy, the energy expended in driving the robot motors is almost twice the energy consumed by the DeepCPG policy. Additionally, it can be observed that the trajectories produced by DeepCPG are smoother (see Fig. 9-(b)). In the case of the real robots, this could also contribute to robot safety since DeepCPG policies would not perform any unbounded actions that could damage the robot actuators.

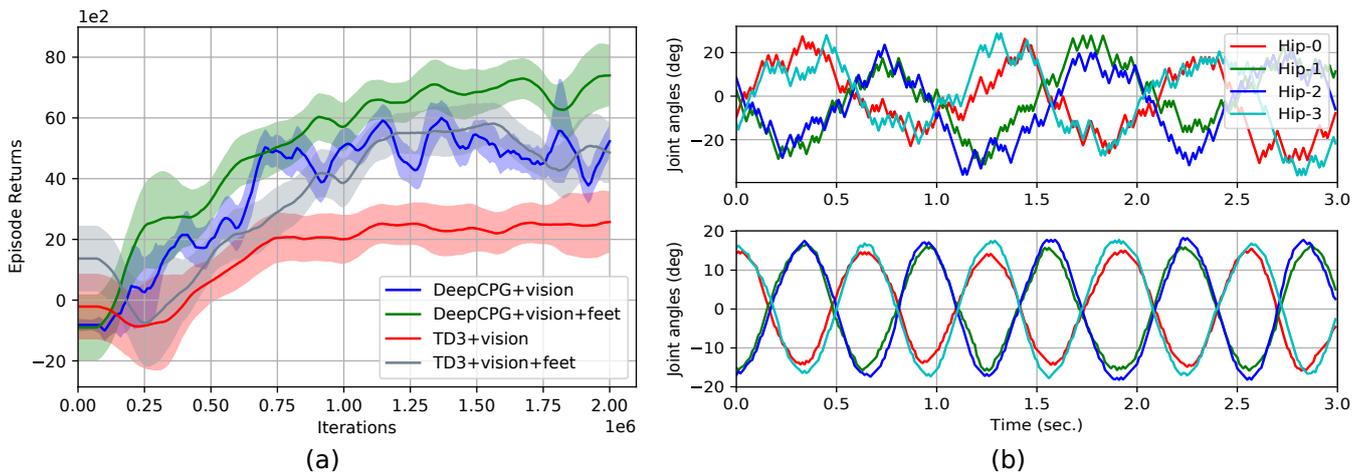

Fig. 9. (a) Training plots for the tasks with visual feedback: (vision) Robot has access to only visual feedback, (vision+feet) Robot gets feedback from vision and feet touch sensors, (TD3) Feed-Forward policy trained using TD3, (DeepCPG) DeepCPG policy; (b) Trajectory of the Hip Joints of the quadruped robot: (Top) Trajectories observed for the Feed-forward MLP policy, (Bottom) Trajectories observed for the DeepCPG policy. Hip-0 is Front Right, Hip-1 is Front Left, Hip-2 is Back Right and Hip-3 is Back left.

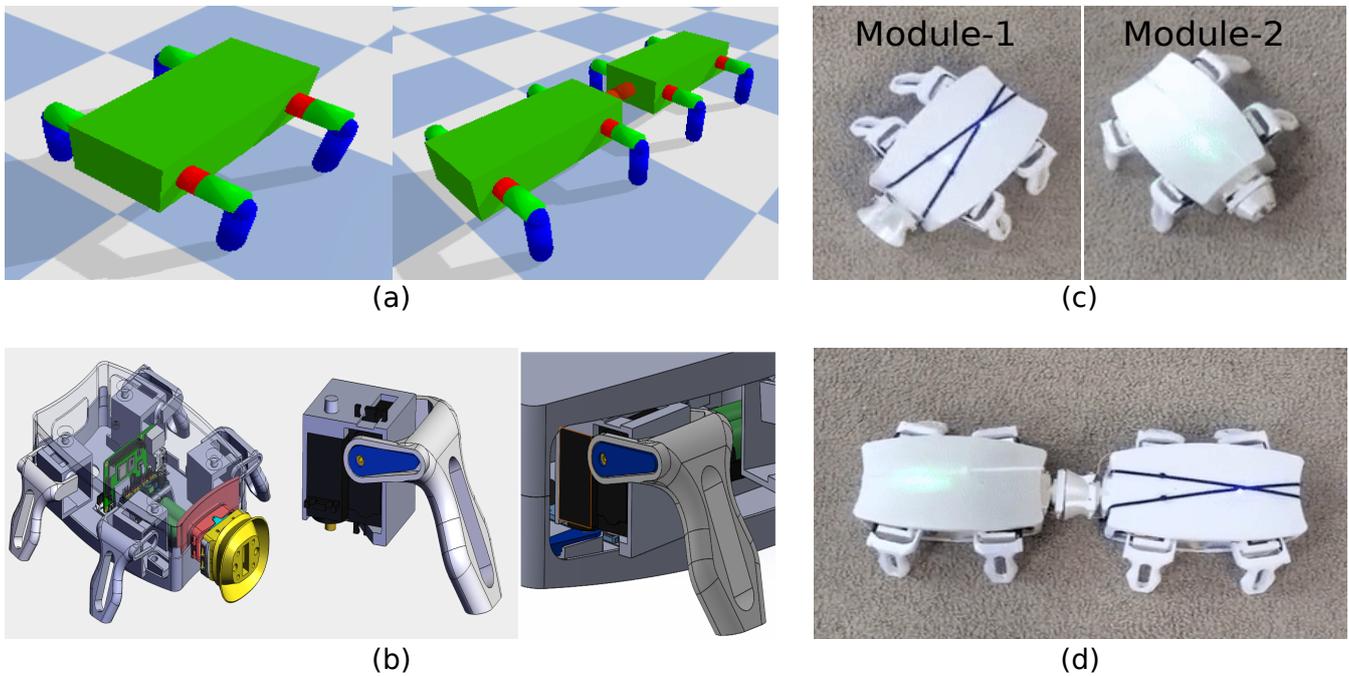

Fig. 10. (a) Physics engine based model: (left) a single robot, (right) a robot with two modules; (b) CAD model: (left) Robot used in Sim2Real experiment, (middle) Single leg along with servos, (right) Leg attached to robot body; (c) Individual modules of real robot; (d) Connected modules of real robot.

### H. Scalability and Simulation to Real-World Transfer of DeepCPG Policies

In this section, results of the scalability study and simulation of the real-world (sim2real) transfer of the DeepCPG policies are presented. Figure 10 shows the snapshots of the simulation model and the corresponding experimental setup used for the sim2real experiments.

The images of the CAD model of the robot used in the simulation-to-real-world transfer experiments with the trained DeepCPG policies are shown in Fig. 10-(b). These robots are custom built and the experiments were performed under a motion capture system. Each of these quadrupeds had two motor joints per leg. Thus, a single quadruped module have a total of 8 degrees of freedom. Raspberry-Pi 4 Model B is used as the on-board computing platform for these robots [80]. The robot joints are connected to a PWM breakout board that handles sending commands to multiple servo motors simultaneously. This is all housed within a shell that gives the robot a turtle-like appearance. These robots can connect and disconnect with each other using an electromagnet. A hall effect sensor is used to detect the connection of one module to another. The connectors are designed with a conical shape to allow for tolerance while connecting (highlighted in yellow in Fig. 10-(b)-(left)). The corresponding 3D-Printed quadruped robot modules using this CAD model are shown in Fig. 10-(c) and Fig. 10-(d).

For the DeepCPG policies trained for sim2real experiments, we did not use touch sensors on the robot feet. The feedback available to the robot consisted of the robot joint angles, joint angle velocities, linear velocity and angular velocity. The inter-modular information shared across each module consisted of values of joint angles measured for each module. As there are eight joints on each module, this information consists of eight joint angle values. The global contextual information shared with Module-2 from Module-1 consisted of robot velocity direction in body frame of reference.

The training plots for the modular system in Fig. 10-(a)-(right) with 16 degrees of freedom are provided in Fig. 11-(a). The policies were trained using the reward function in Eq.(13). This plot compares three training routines. It should be noted that, for a fair comparison, the architectures of the control policies were kept identical in each training routine and only the neural network $\pi_{\theta_m} \forall m \in \{1, 2\}$ parameter initialization strategies were varied:

**Routine-1**: Training the system with Module-1 and Module-2 from scratch where the modular policies were initialized randomly and trained together as a single giant policy. This is referred to as "Not-Mod" in the Fig.11-(a).

**Routine-2**: First, a policy for Module-1 was trained using the reward function Eq. (13) and then transferred to the corresponding module in the new system formed by connecting Module-1 and Module-2. The Module-2 policy weights were initialized randomly. The multi-agent RL setup described in Section V was used to train these control policies. This is referred to as "Mod-Rand" in the Fig.11-(a).

**Routine-3**: First, a policy for Module-1 was trained using the reward function Eq. (13) and then it was used to initialize the weights of each module in the new system formed by connecting Module-1 and Module-2. The multi-agent RL setup described in Section V was used to train these control policies. This is referred to as "Mod" in the Fig.11-(a).

Based on the comparison of all these training routines, it was observed that **Routine-3** is the most sample efficient. In this routine, the modular policies were able to take advantage of two priors when the transfer from an individual module stage to a connected module stage occured. The first prior was as a result of movement primitives embedded in DeepCPG policies. The second prior is due to trained quadruped policy used for parameter initialization in the design complexification stage. It can be observed clearly that modular design complexification with transfer of weights and additive learning enabled effective and efficient scaling of the DeepCPG policies to robots of increasing complexity.

Given the Kuramoto model of CPG embedded in the network, the DeepCPG policies were able to successfully transfer to the real robot without requiring any further fine-tuning of the neural network weights. We performed two experiments with the provided setup:

**Experiment 1** - A DeepCPG policy was trained for a single quadruped robot for waypoint navigation using the reward function in Eq.(12). The policy was trained in simulation and transferred to the real robot. The trajectory followed by the real robot Module-1 is shown in Fig. 11-(b).

**Experiment 2** - In this experiment, the robot Module-1 and Module-2 were connected together. The DeepCPG policy was trained to let the connected modules walk in a straight line. The connection between Module-2 and Module-1 is detected by a hall-effect sensor. When the connection is detected, the modular policy trained using **Routine-3** described above was activated.

Based on these experiments, it was also observed that the learned control policies both for individual robot and the modular robots were able to successfully work in the real world. Additionally, the modular policies also enabled smooth switching of behavior from a single module to a system with two modules and effectively synchronize the leg movements after the connection as decribed in **Experiment 2**. DeepCPG policy also ensured smooth joint trajectories in the robots. This implicitly constrains the robots to follow a smooth and safe trajectories using its actuators.

## VII. Conclusions and Future Work

We presented a developmental RL-based approach for learning robot locomotion that allows the scalable construction of more complex robots by a process of modular complexification and transfer of prior learning. This approach was tested in various simulated environments as well as on a real-world experimental platform. The results show the advantages offered by the behavioral priors introduced in the actor-networks as central pattern generator models. DeepCPG policies were able to show sample efficient learning of various behavioral policies with different sensor modalities. As demonstrated by the results presented for various sensor modalities, DeepCPG policies enable end-to-end learning even in the case of high-dimensional sensor spaces such as visual feedback.

The hierarchical DeepCPG policy also incorporates ideas from the *dual process theory* of human intelligence [81], [82]. In the DeepCPG policy, the lower level CPG policy $\pi_{CPG}$ can be thought of as $System\ 1$ which is a fast, unconscious, and automatic mode of behavior, while the higher-level policy $\pi_\theta$ can be considered as $System\ 2$ which is a slow, conscious and rule-based model of reasoning developed as a result of knowledge gained by the robot from its surrounding environment. In other words, the DeepCPG actor employs the idea of *thinking fast and slow* while learning a behavioral policy.

Closely related work to the proposed work includes the work by Schilling et al. [18] investigating learning of various decentralized control policies for a fixed robot morphology. Huang et al. [83] also presented a very interesting message-passing approach to train a generalized decentralized control policy for a wide variety of agent morphologies. This work showed that an identical modular network could be shared across various actuators in the robot morphology to generate locomotion behaviors. The results using DeepCPG also align with the conclusions drawn from the studies in [18], [83] that decentralized training and biological inspiration are quite helpful for learning. To summarize, decentralized training of DeepCPG policies corroborates well with the observations from these works that modular policies could be trained faster. In addition to that, our work also showed that the interaction of sensory modalities and movement primitives enables faster learning of robust locomotion for a high degree of freedom system.

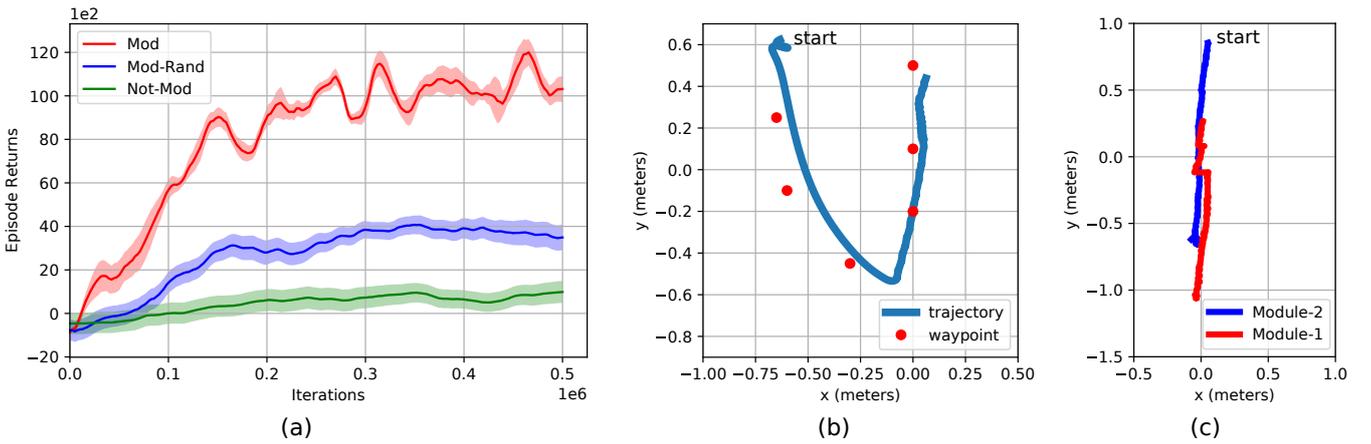

(a)  (b)  (c)

Fig. 11.  (a)Training plots for a modular walking robot when extended from having 4 legs to 8 legs: [Not-Mod] Policy was trained for 8-legged robot from random initialization; [Mod-Rand] Policy was initialized randomly for the added legs when the 4-legged robot was extended to an 8-legged robot; [Mod] Policy was initialized with the weights from the quadruped module when the robot was extended from 4-legged to 8-legged, (b) Trajectory followed by Robot Module-1 for waypoint navigation task, (c) Trajectories followed by Module-1 and Module-2 while walking in the straight line after connecting with each other.

From the perspective of developmental robotics, the approach proposed to train the behavioral policies for modular systems also shows the efficacy of bootstrapping more complex intelligent systems from simpler ones based on biological principles. The gradual complexification with modular policies along with phenotypic development in the robot enables nontrivial sensorimotor integration on a real robotic platform.

We used the Kuramoto model as the basis of CPGs. One can replace this model with other models without the loss of generality. The addition of different nonlinear terms in the CPG models could also enable the emergence of diverse behaviors like jumping with appropriate environmental scenarios. The proposed DeepCPG based policies could also be used with complex robot architectures like humanoid robots [84] and soft robots [85] for learning different behaviors. We believe these could be interesting directions to explore as part of future work.

# APPENDIX: DEEPCPG POLICIES FOR ROBOT LOCOMOTION

**Aditya M. Deshpande, Eric Hurd, Ali A. Minai, Manish Kumar**
University of Cincinnati
Cincinnati, OH 45221
deshpaad@mail.uc.edu; hurdeg@mail.uc.edu; ali.minai@uc.edu; manish.kumar@uc.edu

## A   Derivatives of Kuramoto Model based CPG Network

After incorporating Central Pattern Generators based on Kuramoto Model of oscillations in the neural actor, it remains differentiable. The equations 1-4 describe the model of central pattern generators (CPGs).

$$\dot{\phi}_i = \omega_i + \sum_{j \neq i} a_j w_{ij} \sin(\phi_j - \phi_i - \varphi_{ij}), \tag{1}$$

$$\ddot{a}_i = \alpha_a(\beta_a(A_i - a_i) - \dot{a}_i), \tag{2}$$

$$\ddot{b}_i = \alpha_b(\beta_b(B_i - b_i) - \dot{b}_i), \tag{3}$$

$$y_i = b_i + a_i \sin(\phi_i) \tag{4}$$

Here for a CPG $i$, $\phi_i$ is the rate of change of phase, $\omega_i$ is the natural frequency, $A_i$ is the desired amplitude and $B_i$ is the desired offset. $w_{ij}$ represents the connection weight between the incoming connection from CPG $j$ to CPG $i$. $\varphi_{ij}$ is the connection phase bias between the CPGs $i$ and $j$. $(\alpha_a, \beta_a)$ and $(\alpha_b, \beta_b)$ are the constants. The target trajectory or action command sent to the robot environment by CPG $i$ is given by $y_i$.

We discretize equations 1-4 at time $t$ for analytically solution. Equations 5-6 represent the discrete form of Eq. 1.

$$\dot{\phi}_i^t = \omega_i + \sum_{j \neq i} a_j^{t-1} w_{ij} \sin(\phi_j^{t-1} - \phi_i^{t-1} - \varphi_{ij}), \tag{5}$$

$$\phi_i^t = \phi_i^{t-1} + \dot{\phi}_i^{t-1} \delta t \tag{6}$$

Equations 7-9 represent the discrete form of Eq. 2.

$$\ddot{a}_i^t = \alpha_a(\beta_a(A_i - a_i^{t-1}) - \dot{a}_i^{t-1}), \tag{7}$$

$$\dot{a}_i^t = \dot{a}_i^{t-1} + \ddot{a}_i^{t-1} \delta t, \tag{8}$$

$$a_i^t = a_i^{t-1} + \dot{a}_i^{t-1} \delta t, \tag{9}$$

Equations 10-12 represent the discrete form of Eq. 3.

$$\ddot{b}_i^t = \alpha_b(\beta_b(B_i - b_i^{t-1}) - \dot{b}_i^{t-1}), \tag{10}$$

$$\dot{b}_i^t = \dot{b}_i^{t-1} + \ddot{b}_i^{t-1} \delta t, \tag{11}$$

$$b_i^t = b_i^{t-1} + \dot{b}_i^{t-1} \delta t, \tag{12}$$

Similarly, Eq. 13 represent the discrete form of Eq. 4.

$$y_i^t \;=\; b_i^t + a_i^t \sin(\phi_i^t) \tag{13}$$

Using 6-13, a recurrent relationship can be derived for $y_i$ and its derivatives with respect to CPG parameters $\{\varphi_{ij}, w_{ij}, \omega_i, A_i, B_i\} \forall i, j \in \{1, .., N\}$. This relationship is similar to the one discussed for discrete dynamic movement primitives in [Gams et al., 2018].

Let there be a cost function $\mathcal{L}(y_i^t)$ which is dependent on the CPG output $y_i^t$ at time $t$. This loss function can be considered as a negative reward function. Thus, to backpropagate through this policy using the loss function, we need the partial derivatives with respect to the parameters that are outputs of the neural policy, i.e., CPG parameters $\{\varphi_{ij}, w_{ij}, \omega_i, A_i, B_i\} \forall i, j \in \{1, .., N\}$. We can derive these partial derivatives using the Chain Rule in calculus.

### A.1 Phase Bias $\varphi$

$$\frac{\partial \mathcal{L}(y_i^t)}{\partial \varphi_{ij}} \;=\; \frac{\partial \mathcal{L}(y_i^t)}{\partial y_i^t} \frac{\partial y_i^t}{\partial \phi_i^t} \frac{\partial \phi_i^t}{\partial \varphi_{ij}} \tag{14}$$

Here, the term $\frac{\partial \phi_i^t}{\partial \varphi_{ij}}$ can be found using Eq. 6:

$$\frac{\partial \phi_i^t}{\partial \varphi_{ij}} \;=\; \frac{\partial \phi_i^{t-1}}{\partial \varphi_{ij}} + \frac{\partial \dot{\phi}_i^{t-1}}{\partial \varphi_{ij}} \delta t \tag{15}$$

And using Eq. 5, to evaluate $\frac{\partial \dot{\phi}_i^{t-1}}{\partial \varphi_{ij}}$ as follows:

$$\frac{\partial \dot{\phi}_i^{t-1}}{\partial \varphi_{ij}} \;=\; \frac{\partial \sum_{k \neq i} a_j^{t-2} w_{ik} \sin(\phi_k^{t-2} - \phi_i^{t-2} - \varphi_{ik})}{\partial \varphi_{ij}} \tag{16}$$

$$=\; \frac{\partial \big( \cdots + a_j^{t-2} w_{ij} \sin(\phi_j^{t-2} - \phi_j^{t-2} - \varphi_{ij}) + \cdots \big)}{\partial \varphi_{ij}} \tag{17}$$

$$=\; a_j^{t-2} w_{ij} \Big( \frac{\partial \phi_j^{t-2}}{\partial \varphi_{ij}} - \frac{\partial \phi_i^{t-2}}{\partial \varphi_{ij}} - 1 \Big) \cos(\phi_j^{t-2} - \phi_i^{t-2} - \varphi_{ij}) \tag{18}$$

But, $\phi_j^{t-2}$ depends on $\varphi_{ji}$ and not on $\varphi_{ij}$. We can rewrite Eq. 18 as Eq. 19

$$\frac{\partial \dot{\phi}_i^{t-1}}{\partial \varphi_{ij}} \;=\; a_j^{t-2} w_{ij} \cos(\phi_j^{t-2} - \phi_i^{t-2} - \varphi_{ij}) \Big( -\frac{\partial \phi_i^{t-2}}{\partial \varphi_{ij}} - 1 \Big) \tag{19}$$

By substituting Eq. 19 in Eq. 15

$$\frac{\partial \phi_i^t}{\partial \varphi_{ij}} \;=\; \frac{\partial \phi_i^{t-1}}{\partial \varphi_{ij}} + a_j^{t-2} w_{ij} \cos(\phi_j^{t-2} - \phi_i^{t-2} - \varphi_{ij}) \Big( -\frac{\partial \phi_i^{t-2}}{\partial \varphi_{ij}} - 1 \Big) \delta t \tag{20}$$

The loss function $\mathcal{L}(y_i^t)$ can be given similar treatment with respect to $w_{ij}$ and $\omega_i$.

### A.2 CPG Connection Weights $w$

$$\frac{\partial \phi_i^t}{\partial w_{ij}} \;=\; \frac{\partial \phi_i^{t-1}}{\partial w_{ij}} + a_j^{t-2} \sin(\phi_j^{t-2} - \phi_i^{t-2} - \varphi_{ij}) \delta t + a_j^{t-2} w_{ij} \cos(\phi_j^{t-2} - \phi_i^{t-2} - \varphi_{ij}) \Big( -\frac{\partial \phi_i^{t-2}}{\partial w_{ij}} \Big) \delta t \tag{21}$$

### A.3 CPG Natural Frequency $\omega$

$$\frac{\partial \phi_i^t}{\partial \omega_i} = \frac{\partial \phi_i^{t-1}}{\partial \omega_i} + \delta t + a_j^{t-2} w_{ij} \sin(\phi_j^{t-2} - \phi_i^{t-2} - \varphi_{ij})\Big(-\frac{\partial \phi_i^{t-2}}{\partial \omega_i}\Big)\delta t \tag{22}$$

### A.4 CPG Desired Amplitude $A$

Similarly, taking partial derivative of loss function $\mathcal{L}(y_i^t)$ with respect to $A_i$:

$$\frac{\partial \mathcal{L}(y_i^t)}{\partial A_i} = \frac{\partial \mathcal{L}(y_i^t)}{\partial y_i^t}\frac{\partial y_i^t}{\partial a_i^t}\frac{\partial a_i^t}{\partial A_i} \tag{23}$$

Using Eq. 9 to find the value of $\frac{\partial a_i^t}{\partial A_i}$ in Eq. 23:

$$\frac{\partial a_i^t}{\partial A_i} = \frac{\partial a_i^{t-1}}{\partial A_i} + \frac{\partial \dot{a}_i^{t-1}}{\partial A_i}\delta t \tag{24}$$

Using Eq. 8 we can evaluate $\frac{\partial \dot{a}_i^{t-1}}{\partial A_i}$:

$$\frac{\partial \dot{a}_i^{t-1}}{\partial A_i} = \frac{\partial \dot{a}_i^{t-2}}{\partial A_i} + \frac{\partial \ddot{a}_i^{t-2}}{\partial A_i}\delta t \tag{25}$$

Similarly, Eq. 7 to find $\frac{\partial \ddot{a}_i^{t-1}}{\partial A_i}$:

$$\frac{\partial \ddot{a}_i^{t-1}}{\partial A_i} = \alpha_a\Big(\beta_a(1 - \frac{\partial a_i^{t-2}}{\partial A_i}) - \frac{\partial \dot{a}_i^{t-2}}{\partial A_i}\Big) \tag{26}$$

By substitution for Eq. 25 in Eq. 24,

$$\frac{\partial a_i^t}{\partial A_i} = \frac{\partial a_i^{t-1}}{\partial A_i} + \Big(\frac{\partial \dot{a}_i^{t-2}}{\partial A_i} + \frac{\partial \ddot{a}_i^{t-2}}{\partial A_i}\delta t\Big)\delta t \tag{27}$$

By substitution Eq. 26 in Eq. 27,

$$\frac{\partial a_i^t}{\partial A_i} = \frac{\partial a_i^{t-1}}{\partial A_i} + \Big(\frac{\partial \dot{a}_i^{t-2}}{\partial A_i} + \alpha_a\big(\beta_a(1 - \frac{\partial a_i^{t-3}}{\partial A_i}) - \frac{\partial \dot{a}_i^{t-3}}{\partial A_i}\big)\delta t\Big)\delta t \tag{28}$$

### A.5 CPG Desired Offset $B$

Now, taking partial derivative of loss function $\mathcal{L}(y_i^t)$ with respect to $B_i$ and again using the Chain Rule:

$$\frac{\partial \mathcal{L}(y_i^t)}{\partial B_i} = \frac{\partial \mathcal{L}(y_i^t)}{\partial y_i^t}\frac{\partial y_i^t}{\partial b_i^t}\frac{\partial B_i^t}{\partial B_i} \tag{29}$$

Now, we can derive the similar expression as Eq. 28 for $B_i$ in Eq. 30,

$$\frac{\partial b_i^t}{\partial B_i} = \frac{\partial b_i^{t-1}}{\partial B_i} + \Big(\frac{\partial \dot{b}_i^{t-2}}{\partial B_i} + \alpha_b\big(\beta_b(1 - \frac{\partial b_i^{t-3}}{\partial B_i}) - \frac{\partial \dot{b}_i^{t-3}}{\partial B_i}\big)\delta t\Big)\delta t \tag{30}$$

Therefore, equations 20, 21, 22, 28, 30 show that the dynamical system defined by Central Pattern Generators based on Kuramoto model of oscillation is differentiable with respect to $\varphi_{ij}, w_{ij}, \omega_i, A_i, B_i$ respectively.

## B Pseudo code

---

**Algorithm 1:** Training DeepCPG Policies

---

**Input:** Maximum iterations $Nmax$, Babbling steps $\tau_b$, Update steps $\tau_{update}$, Policy update delay $\tau_{delay}$, CPG
      policy steps $\tau_c$, Observation steps $\tau_o$, Discount factor $\gamma$, Polyak averaging constant $\rho$, Learning rate $\eta$,
      Noise clip limit $c$, Policy exploration noise $\sigma$

**NOTE**: We use $x_{-\tau_{(\cdot)}:} \equiv x_{t-\tau_{(\cdot)}:t}$ and $x_{:\tau_{(\cdot)}} \equiv x_{t:t+\tau_{(\cdot)}}$ to avoid notation clutter

Randomly initialize $\pi_\theta$, Q-functions $Q_{\kappa_1}$ and $Q_{\kappa_2}$

Set target network parameters equal to main parameters: $\theta^{targ} \leftarrow \theta$, $\kappa_1^{targ} \leftarrow \kappa_1$ and $\kappa_2^{targ} \leftarrow \kappa_2$

Initialize replay buffer $\mathcal{R}$

**while** $k < Nmax$ **do**

    Observe state $s_{-\tau_o:}$, Goal $g$

    **if** $k < \tau_b$ **then**

        |  Select CPG goals $g_{CPG}$ randomly

    **else**

        |  Select CPG goals $g_{CPG} \sim \pi_\theta(s_{-\tau_o:}, g)$

    **end**

    **for** $n = \{1, \ldots, \tau_c\}$ **do**

        Observe CPG hidden state $h$

        Select CPG actions $\hat{g}_j, h' \sim \pi_{CPG}(g_{CPG}, h)$

        $g_j = clip\,(\hat{g}_j + \epsilon, g_j^{\;min}, g_j^{\;max})$ where $\epsilon \sim \mathcal{N}(0, \sigma)$; Execute $g_j$ in environment

        Observe next state $s'$, reward $r$ and environment terminal signal $d$ indicating if $s'$ is terminal state

        Collect $(s, g_j, s', r, d, h, h', g_{CPG}, g)$ in $\mathcal{R}$ where $s = s_t$

        $k = k + 1$

    **end**

    If $s'$ is terminal state, reset environment, reset goal $g$ and reset CPG hidden state $h$

    **if** $k > \tau_b$ **then**

        **for** $i = \{1, \ldots, \tau_{update}\}$ **do**

            Randomly sample batches from episodes stored in $\mathcal{R}$

            $B = \{(s_{-\tau_o:}, g_{j,:\tau_c}, s'_{:\tau_c}, r_{:\tau_c}, d_{t+\tau_c}, h_t, h'_{t+\tau_c}, g_{CPG_t}, g_t, g_{t+\tau_c})\}$

            Compute $g'_{CPG} \sim \pi_{\theta^{targ}}(s'_{t+\tau_c-\tau_o:t+\tau_c}, g_{t+\tau_c})$

            $h' = h'_{t+\tau_c}$

            **for** $t = \{1, \ldots, \tau_c\}$ **do**

                Get CPG hidden state $h'$

                Select target CPG actions $\hat{g}'_{j,t}, h'' \sim \pi_{CPG}(g'_{CPG}, h')$

                Collect $g'_{j,t} = clip\,(\hat{g}'_{j,t} + clip\,(\epsilon, -c, c)\,, g_j^{\;min}, g_j^{\;max})$ where $\epsilon \sim \mathcal{N}(0, \sigma)$

            **end**

            Compute targets:

            $y(r_{:\tau_c}, s'_{t+\tau_c-\tau_o:t+\tau_c}, g_{t+\tau_c}, d_{t+\tau_c}) =$

            $\sum r_{:\tau_c} + \gamma(1 - d_{t+\tau_c}) \min\limits_{i=1,2} \left( Q_{\kappa_i^{targ}}\left(s'_{t+\tau_c-\tau_o:t+\tau_c}, g_{t+\tau_c}, g'_{j,:\tau_c}\right) \right)$

            Update Q-functions by one-step gradient descent using:

            $\nabla_{\kappa_i} \frac{1}{|B|} \sum\limits_{B} \left( Q_{\kappa_i}(s_{-\tau_o:}, g_t, g_{j,:\tau_c}) - y(r_{:\tau_c}, s'_{t+\tau_c-\tau_o:t+\tau_c}, g_{t+\tau_c}, d_{t+\tau_c}) \right)^2$   for $i = \{1, 2\}$

            **if** $i \mod \tau_{delay} = 0$ **then**

                Update policy by one-step gradient ascent $\nabla_\theta \frac{1}{|B|} \sum\limits_{B} Q_{\kappa_1}(s_{-\tau_o:}, g_t, g_{j,:\tau_c})$

                Update target networks:

                $\theta^{targ} \leftarrow \rho\theta^{targ} + (1 - \rho)\theta;$    $\kappa_i^{targ} \leftarrow \rho\kappa_i^{targ} + (1 - \rho)\kappa_i$ for $i = \{1, 2\}$

            **end**

        **end**

    **end**

**end**

---

**Algorithm 2:** Deployment of trained DeepCPG Policy in Robot Environment

---

**Input:** Trained policy $\pi_\theta$, CPG policy steps $\tau_c$, Observation steps $\tau_o$, Goal $g$
**NOTE:** We use $x_{-\tau_{(\cdot)}:} \equiv x_{t-\tau_{(\cdot)}:t}$ and $x_{:\tau_{(\cdot)}} \equiv x_{t:t+\tau_{(\cdot)}}$ to avoid notation clutter
Initialize CPG hidden state $h$
**while** *Robot is active* **do**
    Observe state $s_{-\tau_o:}$
    Select CPG goals $g_{CPG} \sim \pi_\theta(s_{-\tau_o:}; g)$
    **for** $n = \{1, \ldots, \tau_c\}$ **do**
        Observe CPG hidden state $h$
        Select CPG actions $g_j, h' \sim \pi_{CPG}(g_{CPG}, h)$
        Execute $g_j$ in environment
        Observe next state $s'$, reward $r$
    **end**
**end**

---

## C   Schematic of DeepCPG Policy

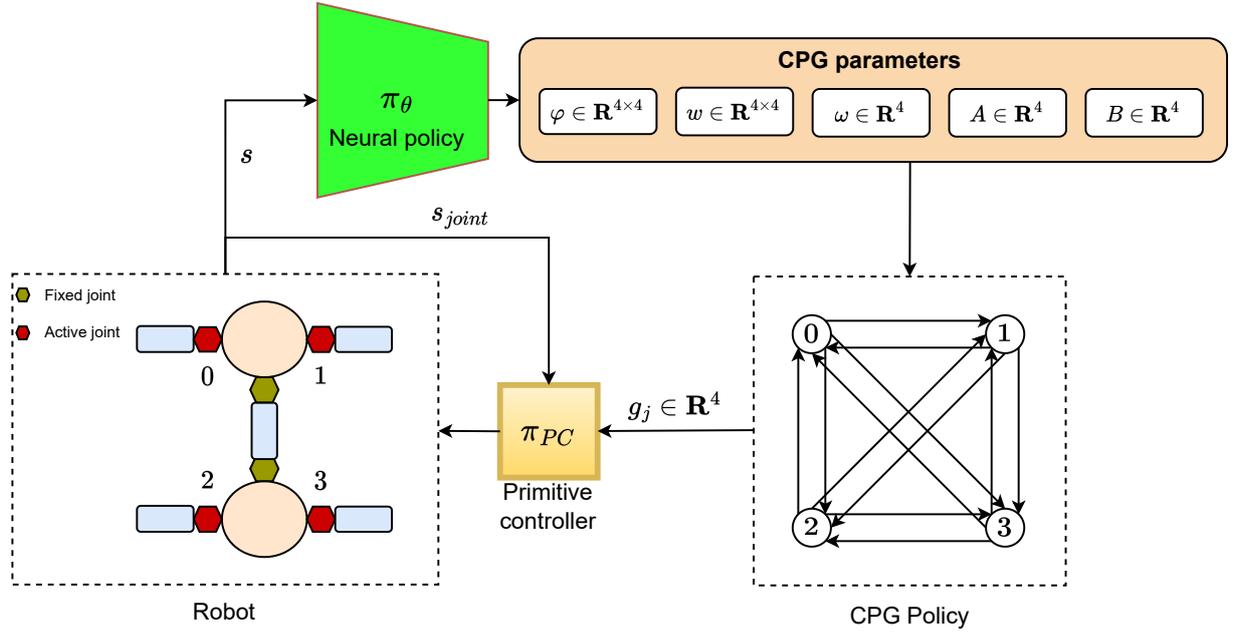

Figure 1: Example schematic illustrating CPG Policy and dimensions of CPG parameter outputs from neural policy $\pi_\theta$ for a robot with four CPG nodes. In CPG Policy, each circled number represents a CPG node corresponding to the active joint on the robot with that number. It should be noted that this is just an illustration. In the case of the actual robots considered for the experiments, the number of joints on the robot is more and the corresponding diagram of CPG Policy would have far more connections. We chose to show an example with only four nodes in the CPG policy to avoid clutter. Please refer Figure 2 and Section-III in the paper for the meaning of various symbols used here.

## D    Additional Results

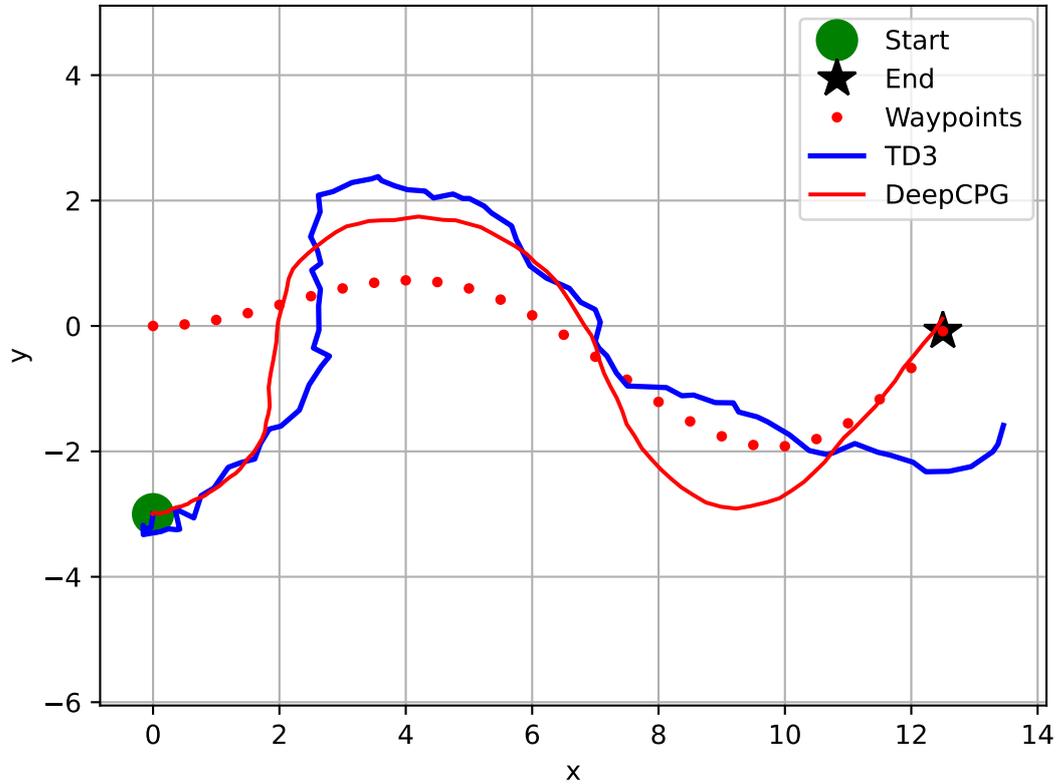

Figure 2: Here we show the trajectories followed by trained policies for the go-to-goal task. In both cases, the robot starts in an identical state. The target waypoints are shown to the robot using a receding horizon strategy. Section VI-C of the paper provides further details of this task. The episode length was 3000 timesteps. A new waypoint was introduced every 100 timesteps and a total of 26 waypoints were given to the robot. Legends: (Start) Start location of the robot, (End) Final target waypoint, (Waypoints) True target waypoints given to the robot, (TD3) Path traced by Feed-forward policy trained using TD3, (DeepCPG): Path traced by trained DeepCPG policy.